\documentclass[pdflatex,sn-mathphys]{sn-jnl}

\jyear{2021}%

\usepackage[caption=false]{subfig}
\usepackage{xcolor}
\usepackage{colortbl}
\usepackage{enumitem}
\usepackage[export]{adjustbox}
\usepackage{comment}
\usepackage{graphicx}

\theoremstyle{thmstyleone}%
\theoremstyle{thmstyletwo}%

\theoremstyle{thmstylethree}%

\begin{document}

\title[WaRTEm-AD]{Warping Resilient Scalable Anomaly \\ Detection in Time Series}


\author*[1]{\fnm{Abilasha} \sur{S}}\email{111814001@smail.iitpkd.ac.in}

\author[1]{\fnm{Sahely} \sur{Bhadra}}\email{sahely@iitpkd.ac.in}

\author[2]{\fnm{Deepak} \sur{P}}\email{deepaksp@acm.org}

\author[1]{\fnm{Anish} \sur{Mathew}}
\email{anishmm1997@gmail.com}

\affil*[1]{\orgdiv{Computer Science and Engineering}, \orgname{IIT Palakkad}, \orgaddress{\street{Kanjikode}, \city{Palakkad}, 
\state{Kerala}, \country{India}}}

\affil[2]{\orgdiv{School of EEECS}, \orgname{Queen's University Belfast}, \orgaddress{\street{University Road}, \city{Belfast}, 
\country{UK}}}


\abstract{Time series data is ubiquitous in the real-world problems across various domains including healthcare, social media, and crime surveillance. Detecting anomalies, or irregular and rare events, in time series data, can enable us to find abnormal events in any natural phenomena, which may require special treatment. Moreover, labeled instances of anomaly are hard to get in time series data. On the other hand, time series data, due to its nature, often exhibits localized expansions and compressions in the time dimension which is called warping. These two challenges make it hard to detect anomalies in time series as often such warpings could get detected as anomalies erroneously. Our objective is to build an anomaly detection model that is robust to such warping variations. In this paper, we propose a novel unsupervised time series anomaly detection method, WaRTEm-AD, that operates in two stages. Within the key stage of representation learning, we employ data augmentation through bespoke time series operators which are passed through a twin autoencoder architecture to learn warping-robust representations for time series data. Second, adaptations of state-of-the-art anomaly detection methods are employed on the learnt representations to identify anomalies. We will illustrate that WaRTEm-AD is designed to detect two types of time series anomalies: point and sequence anomalies. We compare WaRTEm-AD with the state-of-the-art baselines and establish the effectiveness of our method both in terms of anomaly detection performance and computational efficiency.}




\maketitle


\section{Introduction}

With electronic technology and sensors permeating every sphere of life, there are tremendous amounts of digital time series data being generated every moment. All over the world, person-level data measurement devices which generate daily sequences such as medical wearables\footnote{\tiny https://www.businesscloud.co.uk/news/nhs-to-give-thousands-of-free-wearables-to-reduce-diabetes} and activity trackers\footnote{\tiny https://en.wikipedia.org/wiki/Activity\_tracker} are gaining popularity. In this case, the dataset would naturally be a set of day-level time series sequences, and the data object would be an individual time series. On the other hand, population level time series such as sequences of daily footfalls in a hospital or bank may be considered as a single large time series sequence. Each such sequence is itself a dataset comprising the sequence of points within it, with each individual point or a small sub-sequence considered as a data object.  Anomalous temporal patterns over such data could point to emerging epidemics or health issues (in hospital data) or financial instability (or rumours thereof) of financial institutions (in bank data). In the former case of medical wearables, the anomaly is at the level of the daily sequence (we call them as {\it sequence anomalies}), whereas the latter case involves anomalous point-events or sub-sequences (called {\it point anomalies} or {\it sub-sequence anomalies}) within a long time series.  

\begin{wrapfigure}[20]{r}{0.5\linewidth}
        \vspace{-0.3in}
         \centering
         \includegraphics[height=5.3cm, width=0.48\textwidth]{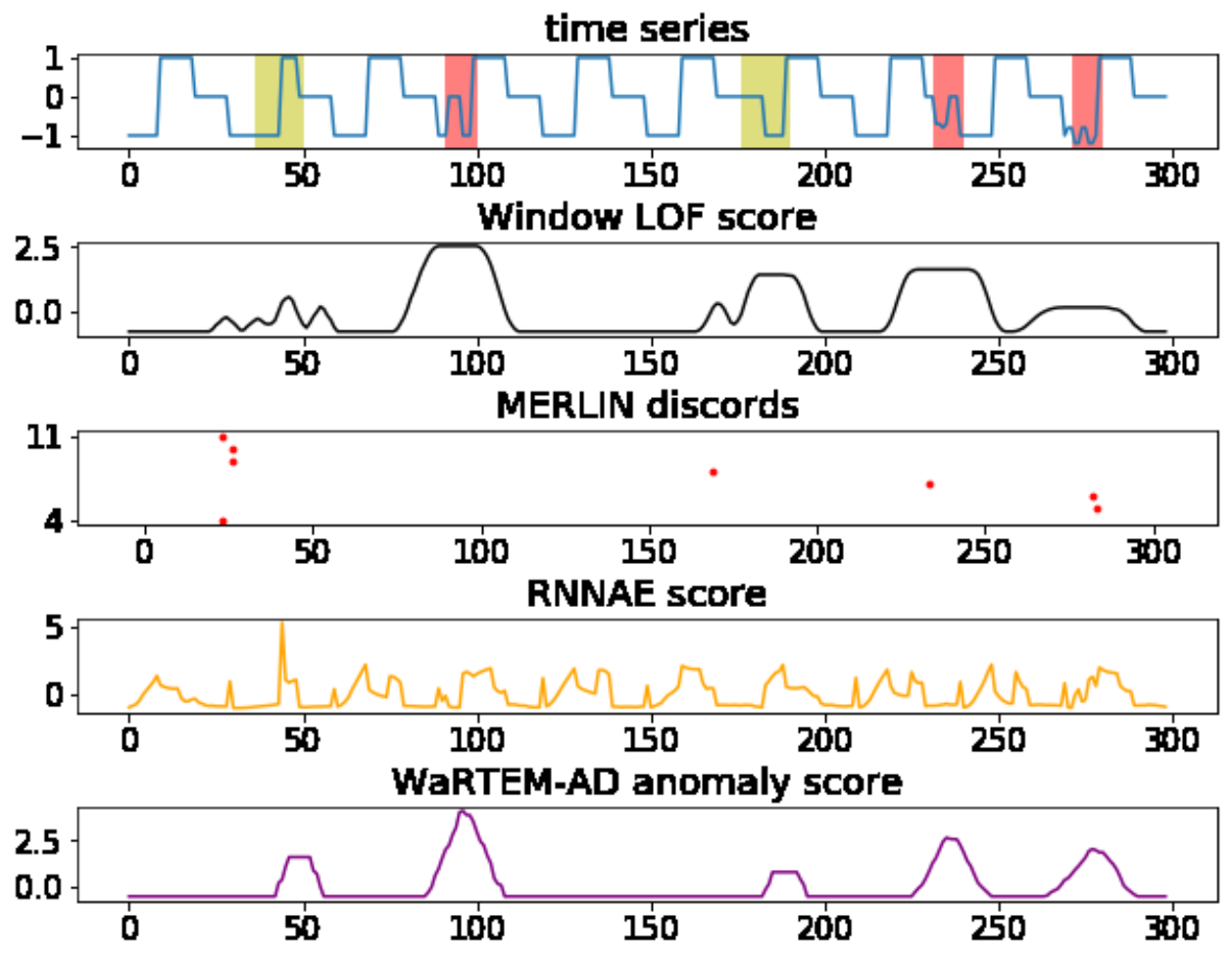}
          \caption{Illustration of point anomalies and its challenges. Top most plot shows time series with point anomalies (red regions), warp variations (green
regions). Lower plots shows anomalies detected (denoted by spikes) by Euclidean distance based LOF
, RNNAE
, MERLIN
and proposed WaRTEm-AD respectively while $x$ axis indicate timestamps.}
\label{fig:performance}
\end{wrapfigure}

\noindent{\bf Time Series Warping:} Time series anomaly detection is a challenging task, because normal and abnormal behaviours depend on the context and various temporal dependencies. Moreover, due to several legitimate reasons, time series data is often locally compressed or expanded keeping the high-level temporal pattern unchanged. These localized expansions and contractions are called {\it warping}. Such warping variations appear in virtually every time series process, and these should not be interpreted as semantic variations. For example, accent differences in speech time series would manifest as lengthening and shortening of words. Time series anomaly detection should be cognizant of and robust to such warping effects and ensure that varying accents not be regarded anomalous. These challenges can be illustrated with a point anomaly detection task on a synthetic time series signal as shown in Fig.~\ref{fig:performance}. Some points or sequences in a time series can be within the range of normal data values but still have anomalous behaviour as shown in red patch near $100^{th}$ time stamp in the top most plot of Fig.~\ref{fig:performance}. At the same time, warp variations as shown in green patches in Fig.~\ref{fig:performance} should not be regarded as anomalous. We consider anomaly as any kind of rare and non-mainstream pattern other than warping distortion\\
\noindent{\bf Data-driven Anomaly Detection:} In general, anomaly detection is modelled as an unsupervised task~\citep{Zimek2017,greff2017lstm,ma2019learning}. Such models employ a statistical model for {\it dataset}-level patterns (e.g., clustering or representation learning, as will be seen in Section~\ref{sec:relwork}), followed by assessing individual {\it data objects} within the dataset for conformance to the statistical model. The lower the conformance, the higher the chance of anomalousness. On the other hand, discord discovery based models also find a subsequence of a time series as an anomalous sequence which significantly differs from usual time series patterns~\citep{hotsax,Nakamura2020MERLINPD,series2graph,diskaware,hu2019novel,senin2015time}. \\
\\
\noindent{\bf Existing Methods:} Existing anomaly detection methods are neither designed to handle the presence of warping in the time series datasets specially, nor do so incidentally. For example, Euclidean distance based Local Outlier Factor (LOF) method~\citep{breunig2000lof} (second row of Fig.~\ref{fig:performance}) estimates similarity using Euclidean metric for each sub-sequence and hence consider warping distortions also as anomalous points ($180^{th}$ timestamp). Similar observations hold for the discord based MERLIN~\citep{Nakamura2020MERLINPD} model (third  row of Fig. \ref{fig:performance}) which detects the first warp distortion ($50^{th}$ timestamp) as anomaly. Recent Convolutional Neural Network (CNN) and Recurrent Neural Network (RNN) based models ~\citep{greff2017lstm,ma2019learning} also treat a signal and its warp variants  as different signals and hence tend to detect warping as anomaly. Moreover, deep learning fails to model the extreme values for the sake of generalisation\citep{ding19} and hence it is observed that  the state-of-the-art auto-encoder based vector representation model Recurrent Neural Network Autoencoder Ensembles (RNNAE)~\citep{kieu2019outlier} (third row of Fig. \ref{fig:performance}) assign high anomaly scores to all extreme value points  and hence fails to capture actual anomalous points ($250^{th}$ timestamp). Detecting { warp} variations as anomalies can cause an abundance of false positives and reduce the utility of anomaly detection methods with significant implications in domains like motion capture, gesture recognition, digital signature verification, robotics and astronomy. Therefore, a notion of (dis)similarity between time series that is robust to warping variations, is indispensable in anomaly identification. Dynamic Time  Warping (DTW)~\citep{berndt1994using} has been among the most successful~\citep{mueen2016extracting} warping robust distance measures. But, time complexity for calculating DTW is in the quadratic order of the length of a time series  which makes it non-scalable and hard to be used in anomaly detection over large datasets.\\
\\
\noindent{\bf Our Contribution:} In this paper, we design a novel auto-encoder based deep learning model to achieve robustness against warping in time series anomaly detection. At a very high level, the idea is to obtain warping robust embeddings of time series in a scalable manner through data augmentation and self-supervised learning. We illustrate warping robustness within the learnt space through a local density based anomaly detection method over the embeddings. Our empirical analysis of the effectiveness of the anomaly detection illustrates the effectiveness of our formulation. Our contributions are: 
\begin{itemize}
\item {\bf Novel Direction:} In a first-of-a-kind exploration (to our best knowledge), we consider warping-robustness, a critical feature, in unsupervised time series anomaly identification.
\item {\bf Method:} We propose a warping-robust two-phase anomaly detection framework {\it Warp Resilient Time-series Embedding - Anomaly Detection} (WaRTEm-AD), that can detect both point as well as sequence anomalies:
\begin{itemize}
    \item The first phase, {\it WaRTEm}, involves learning a warping-robust embedding for time series through a novel twin auto-encoder architecture which employs a unique mechanism called {\it warping operators} for data augmentation. 
    \item The second phase makes use of {\it WaRTEm} representations within a local neighborhood-based anomaly detection framework to score points or sequences for anomalousness.
\end{itemize}
\item {\bf Empirical Study:} Through an extensive empirical evaluation over a vast number of real-world datasets and state-of-the-art baselines, we illustrate the effectiveness of WaRTEm-AD in detecting anomalies of varying length and nature. 
\end{itemize}

\section{ Related Work}\label{sec:relwork}
We now summarize related work on {\it point anomalies} and {\it sequence anomalies} separately. 
\vspace{-0.3cm}
\subsection{Point Anomalies} 
There are several families of point anomaly detection approaches based on the type of techniques they use. (I). {\bf Density-based Modelling} defines anomalousness of a point as directly related to the sparsity of the local neighborhood around it. Approaches in this family, viz., LOF~\citep{breunig2000lof,ester1996density} and LOOP~\citep{loop}, consider points as independent, and are thus inherently incapable of modelling temporal dependencies.   {(II)}. {\bf Prediction-error based Approaches}, such as Numenta and NumentaHTM~\citep{lavin2015evaluating,ahmad2017unsupervised} model sequence information using hierarchical temporal memory (HTM) and quantify anomalousness as the difference between the predicted and actual values. {(III)}. {\bf Deep-learning Methods}, while a recent entrant, have demonstrated significant success in the task. While LSTM-AD~\citep{lstmad} uses normal sequences to train a LSTM model, DeepAnT~\citep{munir2018deepant,kieu2018outlier} uses LSTMs and CNNs in a prediction-based formulation. Ensembles of auto-encoders~\citep{kieu2019outlier,chen2017outlier} have been used in a reconstruction-error based scheme, where anomalousness is quantified as directly related to reconstruction error. {(IV)} {\bf Discord Discovery}~\citep{hotsax,Nakamura2020MERLINPD,diskaware,hu2019novel,senin2015time,series2graph,graphan,yeh2016matrix} focuses on finding discords, i.e., time series subsequences that vary significantly from the {\it `usual'} pattern. The top discord, or top-$k$ discords~\citep{diskaware}, are then identified; this scheme is often sensitive to parameters. MERLIN ~\citep{Nakamura2020MERLINPD} is a recent {\it `parameter-free'} discord discovery method and only needs a range of subsequence length (MinL-MaxL); however, the performance of the method is highly sensitive to the setting of these bounds. When discord discovery is adapted to identify points rather than subsequences (e.g., by changing MERLIN's bounds to unity), it morphs to a point anomaly detection method. 
{(V)}. {\bf Multivariate Approaches} exploit correlations~\citep{zhang2019deep} or other relationships~\citep{su2019robust} among different time series for multi variate data to detect anomalies. They are not applicable for univarite time series, the task we address in this paper. Accordingly, We compare our model with LOF, NumentaHTM, MERLIN and RNN autoencoder ensembles(RNNAE)~\citep{kieu2019outlier} which are the state-of-art of methods from each relevant category above.

\subsection{Sequence Anomalies}\label{sec:seqanomaly}
Here the task is to look at the sub-sequences of a long time series and to identify anomalous sub-sequences. The families of techniques are as follows. (I).  {\bf Deep-learning Methods} have been used to model long-range sequential dependencies \citep{greff2017lstm} and cluster structural information \citep{ma2019learning}, followed by using reconstruction error as an indication of anomalousness. BeatGAN~\citep{beatgan} combines Auotencoders and Generative Adversarial Networks trained on non-anomalous datasets while using warp variations to augment the data. (II). {\bf Similarity-based Approaches} target to embed warping resilience in the similarity quantification, DTW~\citep{berndt1994using} being the most successful~\citep{mueen2016extracting,bagnall2017great} among them. It has been also used for anomaly detection~\citep{benkabou2018unsupervised,benkabou2017local} using clustering. However, DTW has a quadratic complexity over both the number of sequences and length of sequences, making it infeasible for large datasets. A recent work~\citep{lei2017similarity} proposed learning a low-dimensional embedding to approximate actual DTW distances using factorization of an $n \times n$ matrix of DTW distances, approximated by using $n \times log(n)$ DTW measurements for efficiency. With each DTW computation being quadratic in sequence length, this is still inefficient. We compare our method against BeatGAN~\citep{beatgan} and DTW~\citep{benkabou2018unsupervised}, these models being the most recent and state-of-art in relevant categories. 
\section{Warp Resilient Time-series Embedding based Anomaly Detection (WaRTEm-AD) Method}
\label{sec:model}

The task that we address in this paper is \textit{to detect point and sequence anomalies in univariate time series in unsupervised manner}. Our focus is to ensure cognizance to warping variations and avoid identifying warp distortions as anomalies. We denote a time series of length $t$ as $\mathcal{T} = \{s_1, s_2, \ldots, s_t\}$, a time-ordered sequence of real numbers, i.e. each $s_i \in \mathbb{R}$. The three anomaly detection tasks have slightly varying problem definitions, described below.

For the {\bf sequence anomaly} task, we consider a data set containing $N$ time series sequences, each of length $t$, i.e., $\mathcal{D} = \{\mathcal{T}_1,\mathcal{T}_2,\ldots ,\mathcal{T}_N\}$ where $\mathcal{T}_i \in \mathcal{D}$ is denoted as $\mathcal{T}_i = \{s_{i1}, s_{i2}, \ldots, s_{it}\}$. The task of sequence anomaly detection, thus, is to find semantically anomalous time series sequences within $\mathcal{D}$. The anomaly detection method involves associating each $\mathcal{T}_i \in \mathcal{D}$ with an anomaly score $AS(\mathcal{T}_i)$; the time series sequences with high anomaly score would be regarded as anomalous. 

For the {\bf point anomaly}  scenario, the dataset comprises of one (long) time series, $\mathcal{T}= \{s_1, s_2, \ldots, s_t\}$, and anomaly detection involves identifying specific data points $s_i$ within $\mathcal{T}$ that may be regarded as non-conformant to the overall pattern(s) within $\mathcal{T}$. The anomaly detection method would associate each data point $s_i \in \mathcal{T}$ with an anomaly score $AS(s_i)$ that quantifies anomalousness. 

For the  {\bf sub-sequence anomaly} case, a dataset consisting of one time series, $\mathcal{T}= \{s_1, s_2, \ldots, s_t\}$, sub-sequence anomaly detection involves identifying specific contiguous sub-sequence of data points within $\mathcal{T}$ that may be regarded as non-conformant to the overall pattern(s) within $\mathcal{T}$. The anomaly detection method would then associate the contiguous sub-sequence $\{s_i,s_{i+1},\ldots,s_{i+k}\}$ with an anomaly score $AS(\{s_i,s_{i+1},\ldots,s_{i+k}\})$. 

\subsection{Warp Resilient Timeseries Embedding (WaRTEm)}

Let us consider the sequence anomaly case first. For $\mathcal{D} = \{\ldots, \mathcal{T}, \ldots \}$ (dropping suffix-representation, i.e., $\mathcal{T}_i$, for convenience), WaRTEm intends to learn warping robust embedding $\mathcal{V} = f(\mathcal{T})$ where $\mathcal{V}\in \mathbb{R}^k$ and $k<m$. The idea is to ensure that the embeddings are robust to warping variations; thus, the embedding of a time series $\mathcal{T}$ and its warped variant $\mathcal{\hat{T}}$ will be placed close by in $\mathbb{R}^k$. Towards learning warping resilient embeddings, we develop novel {\it warping operators} to generate and augment the data with warping variants of sequences in $\mathcal{D}$, inspired by recent learning paradigms such as self-supervision~\citep{kolesnikov2019revisiting}. We first outline the proposed two warping operators (copy and interpolation) and then the {\it WaRTEm} training approach to learn warping resilient embeddings.

\subsubsection{Warping Operators.}
\begin{wrapfigure}[9]{r}{0.45\linewidth}
\begin{center}
\includegraphics[width = 0.5\textwidth]{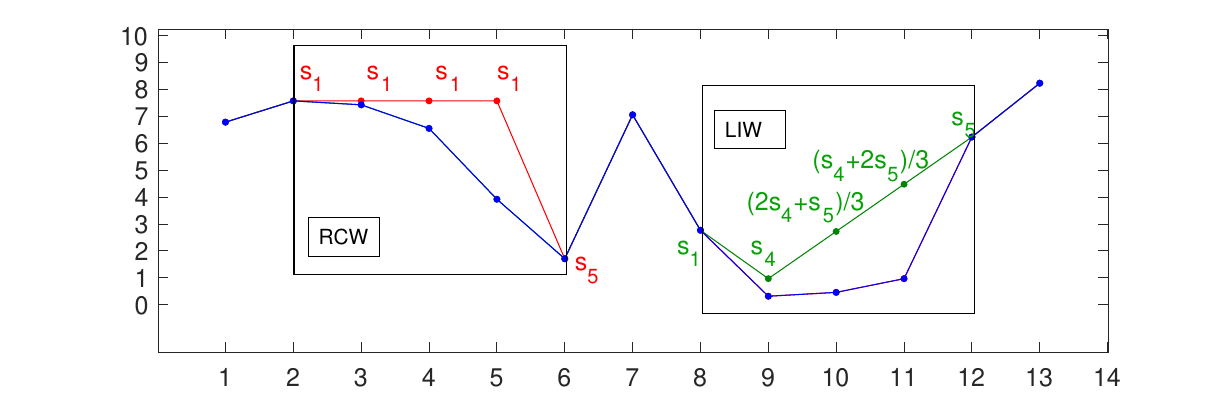}
\caption{Illustration of warping operators. Blue is the original sequence, with red and green indicating RCW and LIW modifications respectively.} 
\label{fig1}
\end{center}
\end{wrapfigure}

A time series $\mathcal{T}$ denoted as $\{s_1 , s_2, \ldots,s_t\}$; may be modified by applying multiple operators upon a chosen contiguous sub-sequence called {\it warping focus window}, as shown in Fig.~\ref{fig1}. The operators are defined below. 

\noindent{\bf Copy Warping: } 
\label{sec:opcw}
 The $LeftCopyWarp$ (LCW) and $RightCopyWarp$(RCW) operator at timestamp $t$ with warping focus window of length $r$ is defined as in Eqn.\ref{eq:RCW}:\\
 In other words, LCW shrinks the left side of the window and extends the right endpoint to a plateau, whereas RCW shrinks the right side of the window and extends the left endpoint to a plateau. It may be noted that a time series $\mathcal{T}$ and its warped variant $LCW(\mathcal{T})$ or $RCW(\mathcal{T})$ differ only in the values within the warping focus window. 

\begin{eqnarray}
LCW_{t,r}(.)&:& [s_{t} , s_{t+1}, \ldots, s_{t+r-2}, s_{t+r-1} ] 
\rightarrow  [s_{t},  s_{t+r-1}, \ldots, s_{t+r-1} , s_{t+r-1} ] \nonumber\\
RCW_{t,r}(.)&:& [s_{t} , s_{t+1}, \ldots, s_{t+r-2}, s_{t+r-1} ] 
\rightarrow  [s_{t},  s_{t}, \ldots, s_{t} , s_{t+r-1} ].  
\label{eq:RCW}
\end{eqnarray}

\noindent{\bf Interpolation Warping:} The left and right interpolation warp operators (LIW and RIW) differ from copy warp in that they fill the deleted sub-sequence using a slope than a plateau. 
\begin{eqnarray}
LIW_{t,r}(.)&:& [s_{t} , s_{t+1}, \ldots, s_{t+r-2}, s_{t+r-1} ]
\rightarrow  [s_{t},  LI(s_{t+r-2},s_{t+r-1},r-1) ] \nonumber \\ 
RIW_{t,r}(.)&:& [s_{t} , s_{t+1}, \ldots, s_{t+r-2}, s_{t+r-1} ]  
\rightarrow  [ LI(s_t, s_{t+1},r-1), s_{t+r-1} ] 
\label{eq:IW}
\end{eqnarray}
where $LI(x,y,l)$ is a $l$-length sequence that linear interpolates from $x$ to $y$. 
\begin{equation}
LI(x,y,l) = [ x, \frac{(l-2) *x + y}{l-1},\frac{(l-3) *x + 2*y}{l-1} \ldots, \frac{x+(l-2)*y}{l-1}, y] \nonumber
\end{equation}
The design of the operators characterizes the kind of differences that could be induced by time series warping, which should be treated as semantically similar to the original, given our intent of treating warping as non-anomalous. Warping operators provide a way to {\it teach} the model as to what kind of time series variations it should develop a blind spot to, so the network capacity can be better focused.

\subsubsection{Twin Auto-Encoder Architecture.}
\label{sec:twinarch}
\begin{figure}[h!]
\begin{center}
\includegraphics[width =\textwidth]{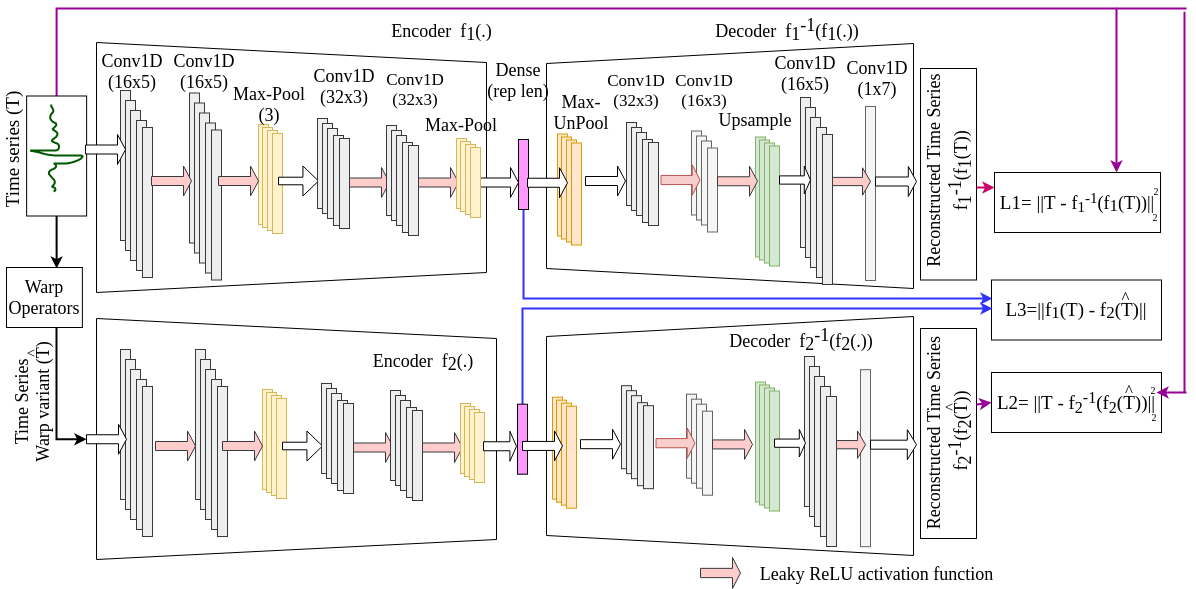}
\caption{WaRTEm Twin Auto-encoder Architecture} \label{fig22}
\end{center}
\vspace{-0.3in}
\end{figure}

The WaRTEm neural network architecture comprising of twin auto-encoders (AEs) is illustrated in Fig.~\ref{fig22}.  Each leg consists of an encoder and a decoder. Encoder are fed with pair inputs $[\mathcal{T},RW(\mathcal{T})]$ or $[LW(\mathcal{T}),\mathcal{T}]$ where $\mathcal{T} \in \mathcal{D}$ with $RW(\mathcal{T})$ and $LW(\mathcal{T})$ indicating warped variants generated using {\it right} and {\it left} warping operators introduced earlier but encoder are expected to learn representation with information in original signal and hence both decoders are expected to reconstruct the unwarped time series $\mathcal{T}$. Notice the left-to-right ordering dependency in the paired input; any left-warped variant appears as the first element, and the right-warped variant appears as the second element.  

Unlike Siamese networks~\citep{chopra2005learning}, the weights are not tied, but the learning is linked together through the loss function L (Eqn.~\ref{eq:loss}). L has three components L1, L2, and L3. L1 is the conventional auto-encoder loss, L2 loss is the  reconstruction loss of non-warp sequence from warp sequence which is similar to the denoising auto-encoder, whereas L3 is the representation coupling loss. Use of correlation based coupling loss for learning common latent representation has been explored before~\citep{andrew2013deep}; we have used Euclidean distance between two presentation as coupling loss steer the learning towards a common representation. The details of the architecture are further illustrated in Fig.~\ref{fig22}. L3 influences only the encoder parts of the AEs, and does not affect the decoders. L3 ensures that the representation of the sequence and warped variants of it are close to each other in the $\mathbb{R}^k$ space. To allow for some learning flexibility, we do not nudge the representations to be identical (as would be the case if the same decoder were attached to both encoders), but instead minimize deviations through the coupling loss. Thus, the optimization objective is:  

\begin{multline}
\hspace{1cm}
\vspace{-0.3in}
 \min _{f_1,f_2}\overbrace{\|f_1^{-1}(f_1(\mathcal{T})) -\mathcal{T}\|^2_2}^\text{L1} + 
\underbrace{\| f_2^{-1}(f_2(\hat{\mathcal{T}})) -{\mathcal{T}}\|^2_2}_\text{L2} +\underbrace{\lambda \|f_1(\mathcal{T}) - f_2(\hat{\mathcal{T}})\|^2_2}_\text{L3} 
\label{eq:loss}
\end{multline}


\subsubsection{Training Strategy and Embedding.}

For each $\mathcal{T} \in \mathcal{D}$, we generate {\it five} warped variants by using a random left-warping operator (LIW or LCW) and another  {\it five} using a random choice of right-warping. The warping focus window is also sampled randomly. Each of these are paired with $\mathcal{T}$ to generate pairs of the form $[LW(\mathcal{T}), \mathcal{T}]$ and $[\mathcal{T}, RW(\mathcal{T})]$. This augmented data-set (of size $10 \times \| \mathcal{D} \|$) is used to train the neural architecture. As the design suggests, the left (right) leg trains for left-warping (right-warping) robustness. While the elements within the training pairs differ only by one warping operator, the warping robustness achieved through the training process will work to ensure that sequences and their warped variants would also lead to similar representations. 
\vspace{-0.3cm}
\subsubsection{Embedding Generation.} Once the training process is complete, each $\mathcal{T}$ is separately passed through the twin encoder and the average of the codes $f_1(\mathcal{T})$ and $f_2(\mathcal{T})$ is used as the embedding for $\mathcal{T}$. 


\subsection{WaRTEm-AD Point Anomaly Scoring}\label{sec:pointscoring}
\label{sec:pas}
The setting of point anomaly involves one time series $\mathcal{T}$. We instantiate WaRTEm by generating a dataset from it using an $m$ length overlapping sliding window that is slid through the entire length of $\mathcal{T}$, to generate a dataset of $m$-length sequences, $\mathcal{D}(\mathcal{T},m)$. We generate a WaRTEm embedding $\mathcal{V}_i$ for each $\mathcal{T}_i \in \mathcal{D}(\mathcal{T},m)$ learnt by treating $\mathcal{D}(\mathcal{T},m)$ as a sequence dataset. For computing anomaly scores $AS(\mathcal{V}_i)$, we use the $K$-$NN$ distance of each embedding, i.e., the distance of the embedding $\mathcal{V}_i$ to its $K^{th}$ nearest neighbor, as an anomaly score. 
The point anomaly task, however, requires us to assign an anomaly score for each data point, and not to a sliding window sequence. Accordingly, we aggregate the $AS(.)$ scores estimated at the sliding window level to the level of each data point $s \in \mathcal{T}$. 


\begin{equation}
AS(s \in \mathcal{T}) = \frac{1}{\|SW_{\mathcal{T},m,p}(s)\|}\sum_{\mathcal{T}_i \in SW_{\mathcal{T},m,p}(s)} AS(\mathcal{V}_i) 
\label{eq:midpoint}
\end{equation}

where $SW_{\mathcal{T},m,p}(s)$ is the set of $m$-width sliding windows in $\mathcal{T}$ of which $s$ is part of middle $p$ points of a sliding window. In other words, the anomaly score for $s$ is the average of the anomaly scores for the embeddings of sliding windows of which $s$ is part of middle $p$ points. 

\subsection{WaRTEm-AD Sequence Anomaly Scoring}
\label{sec:sas}
For sequence anomaly case, after getting embedding for each sequence, the sequence anomaly scores are computed using the $K$-$NN$ distance. Thus, $AS(\mathcal{V}_i)$ is the distance of $\mathcal{V}_i$ to its $K^{th}$ nearest neighbor. 
\section{Experimental Analysis}

We now describe our empirical evaluation of WaRTEm-AD against state-of-the-art baselines. Our evaluation is in three parts: (i) Comparison on Retrieval Accuracy, (ii) WaRTEm-AD's sensitivity to parameters, and (iii) WaRTEm-AD scalability analysis. We set the WaRTEm-AD hyperparameter $\lambda$ to $2$ to balance the two kinds of losses. 

\noindent{\bf Evaluation:} To decide which points in a time series are outliers, one needs to set a threshold. The points whose $AS(.)$ exceed the threshold can be considered as outliers; however, setting such a threshold is a non-trivial task. Hence, we use area under the precision recall curve (PR-AUC) and the ROC curve (ROC-AUC) to capture the trends across varying thresholds~\citep{rocref}. For extreme class imbalance as in our case where anomalies are a small minority, the ROC-AUC may be regarded as too optimistic (see~\cite{gao2021connet}). Therefore, the PR-AUC is more effective than the ROC-AUC in reflecting the quality of the detector in this case \citep{davis2006relationship}. We also compare WaRTEm-AD retrieval using  Reciprocal Rank (RR) \citep{Craswell2009} scores. RR is a popular evaluation metric used in retrieval tasks, which computes the reciprocal rank of first relevant match. 
\subsection{Point Anomaly Evaluation}
\label{pointevaluation}
\noindent{\bf Dataset:}
Real-world univariate time series data sets, i.e., realAWSCloudwatch, realTraffic, realAdExchange, realTweets and realKnownCause from Numenta Anomaly Benchmark (NAB)\footnote{https://github.com/numenta/NAB} repository have been used evaluation of point anomalies. The NAB data sets have 6 to 15 time series sequences and length of these sequences varies from $\approx$ 2K to  16K.  

\noindent{\bf Baseline Methods:} We compare WaRTEm-AD against several methods, outlined below: 
\begin{itemize}
    \item {\bf RNNAE}~\citep{kieu2019outlier}:  
    As described in the paper, we consider 40 seq2seq autoencoder ensembles, with anomaly scores computed as median of reconstruction errors. 
    \item {\bf NumentaHTM}~\citep{ahmad2017unsupervised}, with hyper-parameters as reported in the paper. 
     \item{\bf MERLIN}~\citep{Nakamura2020MERLINPD}: MERLIN outputs the indices of the top discord of various length specified by a range. We set the range to unity as outlined in Sec~\ref{sec:seqanomaly}, to adapt it to the point anomaly task. Since MERLIN returns only the top discord, PR-AUC and ROC-AUC can't be calculated for it.
     \item {\bf Window LOF}: In an adaptation of LOF for our setting, we use the WaRTEm-AD sliding window framework to find sub-sequences of time series, score them using LOF scores, followed by aggregation using Eq.~\ref{eq:midpoint} to obtain point anomaly scores.
\end{itemize}


\noindent{\bf Parameters and Setup:} We set the length of overlapping sliding window, $m$, as $25$, moving one time-step at a time, so that each time point is shared across multiple windows, rendering the design less sensitive to window size. We have experimented with various K values ($K=\{1,3,5\}$) and p values ($p=\{3,5,7\}$). 



 \begin{table}
\caption{ {\bf Point Anomaly Evaluation:} Comparison of PR-AUC, ROC-AUC  and Reciprocal Rank (RR) scores of WaRTEm-AD, MERLIN, RNNAE,  NumentaHTM and Window LOF. Each element of table shows average performance and std over individual time series in each data set for 3 sets of experiments, with varying values of the anomaly neighborhood parameter(except RNNAE and MERLIN). Top score in each row is shown in boldface, and second best is underlined.} 
\label{tab:pointroc}
\begin{center}
\resizebox{0.8\textwidth}{!}{
\begin{tabular}{|l|c|c|c|c|c|}
\hline
\hline
{\bf Dataset} & {\bf WaRTEm-AD} & {\bf MERLIN} &  {\bf RNNAE} & {\bf Num-HTM} & {\bf Win LOF} \\
\hline
\multicolumn{6}{|c|}{\cellcolor{gray!10}{\bf PR-AUC}}\\
\hline
{AdExchange}&\textbf{0.36$\pm$0.17}&\cellcolor[gray]{0.8}&0.19$\pm$0.12&0.05$\pm$0.10& \underline{0.23$\pm$0.10}\\
{Traffic}&\textbf{0.35$\pm$0.25}  &\cellcolor[gray]{0.8}&\underline{0.22$\pm$0.09}&0.07$\pm$0.11&  0.15$\pm$0.07\\
{AWS}& \textbf{0.24$\pm$0.08} & \cellcolor[gray]{0.8}&\textbf{0.24$\pm$0.18}&\underline{0.19$\pm$0.08}& { 0.17$\pm$0.07}\\
 {Tweets}& \underline{0.20$\pm$0.05}&\cellcolor[gray]{0.8} &\textbf{0.21$\pm$0.08}&0.14$\pm$0.03&    0.12$\pm$0.03\\
 {KnownCause}& \textbf{0.23$\pm$0.13}&\cellcolor[gray]{0.8}&{0.17$\pm$0.15}&\underline{0.18$\pm$0.11}&  {0.17$\pm$0.10}\\
 \hline

\multicolumn{6}{|c|}{\cellcolor{gray!10}{\bf ROC-AUC}}\\
\hline
{AdExchange}&\textbf{0.68$\pm$0.21}&\cellcolor[gray]{0.8}&0.50$\pm$0.02&0.50$\pm$0.10& \underline{0.62$\pm$0.16}\\
{Traffic}&\textbf{0.66$\pm$0.20}  &\cellcolor[gray]{0.8}&0.53$\pm$0.03&\underline{0.59$\pm$0.11}&  0.47$\pm$0.11\\
{AWS}& \textbf{0.62$\pm$0.12} & \cellcolor[gray]{0.8}&\underline{0.55$\pm$0.09}&0.47$\pm$0.08 & \underline{0.55$\pm$0.06}\\
 {Tweets}& \textbf{0.60$\pm$0.08}&\cellcolor[gray]{0.8}&\underline{0.58$\pm$0.08}&0.52$\pm$0.10&  0.52$\pm$0.04\\
 {KnownCause}& \textbf{0.62$\pm$0.12}&\cellcolor[gray]{0.8}&0.52$\pm$0.13&\underline{0.54$\pm$0.13} &  0.53$\pm$0.07\\
 \hline
 \multicolumn{6}{|c|}{\cellcolor{gray!10}{\bf RR (Reciprocal Rank) scores}}\\
\hline
 
{AdExchange} &\textbf{0.74$\pm$0.38} & 0.03$\pm$0.06&{0.42$\pm$0.46}&0.49$\pm$0.30&\underline{0.51$\pm$0.24}\\
 {Traffic}& \textbf{0.60$\pm$0.50} & 0.14$\pm$0.38&0.32$\pm$0.47&0.42$\pm$0.30&\underline{0.49$\pm$0.41}\\
{AWS} &\underline{ 0.78$\pm$0.37}& 0.29$\pm$0.47&\textbf{0.80$\pm$0.37}&0.39$\pm$0.28&0.36$\pm$0.42\\
 {Tweets}&0.66$\pm$0.46 & 0.17$\pm$0.41& \textbf{0.92$\pm$0.20}&\underline{0.83$\pm$0.26}&0.29$\pm$0.44\\
 {KnownCause}&\underline{0.35$\pm$0.50}& 0.04$\pm$0.08&\textbf{0.41$\pm$0.46}&0.21$\pm$0.14&\textbf{0.41$\pm$0.48}\\

 \hline
 \hline
\end{tabular}
}
\end{center}
\end{table}

\subsubsection{\bf Results and Analysis}
The first and second one-thirds of the Table~\ref{tab:pointroc} reports the mean PR-AUC and ROC-AUC values. As may be seen therein, WaRTEm-AD outperforms all baselines convincingly. MERLIN cannot be included in the AUC comparisons since the design of the method allows only to identify one top anomaly. 

       \begin{wrapfigure}[19]{r}{0.48\linewidth}
       \vspace{-0.5cm}
         \centering
         \includegraphics[width=0.48\textwidth]{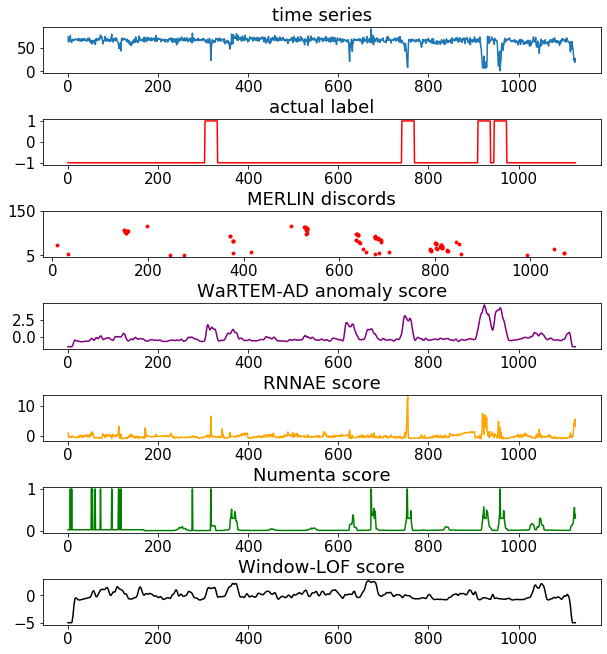}
             \caption{Time series, actual labels \& anomaly scorings of  different methods for  realTraffic (speed\_7578) dataset}
             \label{fig:speednn}
              \end{wrapfigure}
The bottom one-third of Table~\ref{tab:pointroc} reports the RR scores, which can be computed for MERLIN too. It can be observed that WaRTEm-AD method outperforms MERLIN, NumentaHTM and Window LOF with significant margins, whereas RNNAE shows high scores on some datasets; on closer analysis, we found that those datasets contained several high Z-score anomalies, which RNNAE is better at detecting due to its design where the focus is on deviations from sequence-level character (on the other hand, WaRTEm-AD is focused on local neighborhoods). This explains the better performance than WaRTEm-AD on RR too, for those datasets. The comparative AUC results were tested for statistical significance using t-test where it was found that WaRTEm-AD improvements were significant at $p<0.05$, except RNNAE on PR-AUC and RR. 




For qualitative analysis, Fig.~\ref{fig:speednn} presents realTraffic (speed7578) dataset which has 4 anomaly points (around timestamps 300, 700 and 900) and give a visualization of performance of proposed method and all benchmarks. WaRTEm-AD method detects all anomalies without any false positives, whereas MERLIN method fails to detect some anomalies, and incurs several false detections. NumentaHTM scores are high for anomaly points but also detects several normal points as anomalous, whereas Window LOF shows a number of false positives. This also illustrates how RNNAE method is oriented towards identifying high amplitude deviations; given most deviations are downward, observe that the RNNAE score graph appears similar to what a reflection of the time series would look like. 

\subsubsection{\bf Sensitivity Analysis}
\label{sec:analysisp}
\textbf{Analysis over $p$}: The WaRTEm-AD parameter, $p$, controls the aggregation of sub-sequence anomaly scores to the point level. Higher $p$ values reduce the ability to distinguish between adjacent points; for example, with $p=1$, each point gets it's score from a distinct sliding window and with $p=5$, adjacent points share four sliding windows, reducing the difference between the anomaly scores assigned to them.  In our sensitivity analysis, it was observed that the movements over both AUC values are quite smooth with varying values of $p$.  The AUC values were seen to be flatten out for $p$ increasing beyond seven. 

\noindent\textbf{Analysis over $K$}: The performance was also quite stable for variations in $K$; the maximum change in AUC values observed by changing $K$ was $0.04$, indicating stability over $K$ as well. Among the warp operators experimented it is found that interpolation operator showing slightly higher performance than copy operator 
\begin{table}[h!]
\begin{center}
\caption{{\bf Sequence Anomaly Evaluation:} Comparison of PR-AUC, ROC-AUC and RR for WaRTEm-AD, DTW, ANN, BeatGAN (unsupervised version) and LOF. Each element of table shows average performance and std over individual data set for 3 sets of experiments, with varying values of the anomaly neighborhood parameter (except BeGAN-U). Top score in each row is shown in boldface, and second best is underlined} 
\resizebox{0.9\textwidth}{!}{
\begin{tabular}{|l|ccccc|}
\hline
\hline
 \multicolumn{6}{|c|}{\cellcolor{gray!10}{{\bf PR-AUC}}}\\
\hline
Dataset&{WaRTEm-AD}& DTW  & ANN & BeatGAN & LOF \\
&  &  & &\small{(unsupervised)}&\\
\hline
{Toe1}&\textbf{0.55$\pm$0.01}&0.07$\pm$ 0.01&0.14$\pm$ 0.01& 0.17$\pm$ 0.05&\underline{0.44$\pm$ 0.08}\\
{Synthetic}&\underline{0.16$\pm$0.14}&\textbf{0.50$\pm$ 0.01}& {0.14$\pm$ 0.01}& 0.04$\pm$ 0.00&0.06$\pm$ 0.00\\
{Middle}&{0.22$\pm$0.00}&{\bf 0.39$\pm$ 0.00}&0.20$\pm$ 0.14& 0.11$\pm$ 0.00&\underline{0.30$\pm$ 0.03}\\
{HandOut}&\textbf{0.54$\pm$0.00}&\underline{0.47$\pm$ 0.04}& 0.11$\pm$ 0.00& 0.23$\pm$ 0.01& 0.45$\pm$ 0.06\\
{Toe2}&\textbf{0.29$\pm$0.00}&0.06$\pm$ 0.00& 0.14$\pm$ 0.02& 0.13$\pm$ 0.13&\underline{0.20$\pm$ 0.02}\\
{Strawbry}&{0.07$\pm$0.00}&0.62$\pm$ 0.04&\textbf{0.12$\pm$ 0.00}&\underline{ 0.09$\pm$ 0.00}& \underline{0.09$\pm$ 0.00}\\
{ECG5000}&\textbf{0.37$\pm$0.03}&{0.07$\pm$ 0.02}&0.05$\pm$0.00&\underline{0.17$\pm$ 0.00}& 0.09$\pm$ 0.02\\
{Wafer}&\textbf{0.23$\pm$0.00}&0.06$\pm$ 0.00 &\underline{0.11$\pm$0.01}&\underline{0.11$\pm$ 0.01}& 0.09$\pm$ 0.01\\
{Distal}&\underline{0.29$\pm$0.01}&0.20$\pm$ 0.01&0.11$\pm$0.01& 0.21$\pm$ 0.01&\textbf{0.41$\pm$ 0.01}\\
{Proximal}&{0.12$\pm$0.00}&\textbf{0.47$\pm$ 0.01} &0.12$\pm$0.01& 0.06$\pm$ 0.00&\underline{0.29$\pm$ 0.02}\\
{Phalanges}&\underline{0.18$\pm$0.00}& {\bf 0.36$\pm$ 0.01}&0.12$\pm$0.01& 0.11$\pm$ 0.00&\textbf{0.36$\pm$ 0.02}\\
{Dodger}&\textbf{0.66$\pm$0.01}&0.03$\pm$ 0.01&0.21$\pm$0.01& 0.07$\pm$ 0.01& \underline{0.60$\pm$ 0.09}\\
{Earthqks}&{0.06$\pm$0.00}&\textbf{0.17$\pm$ 0.03} &\underline{0.10$\pm$0.02}& 0.08$\pm$ 0.05&{0.09$\pm$ 0.01}\\
{ECG200}&\underline{0.29$\pm$0.01}&{\bf 0.35$\pm$ 0.04}&0.14$\pm$0.04& 0.19$\pm$ 0.04&{0.23$\pm$ 0.06}\\
\hline
\multicolumn{6}{|c|}{\cellcolor{gray!10}{{\bf ROC-AUC}}}\\
\hline 
Dataset&{WaRTEm-AD}& DTW  & ANN & BeatGAN & LOF \\
& &  & &\small{(unsupervised)}&\\
\hline
{Toe1}&\textbf{0.89$\pm$0.01}&0.77$\pm$0.00&0.59$\pm$ 0.03& 0.57$\pm$ 0.05& \underline{0.87$\pm$0.07}\\
{Synthetic}&\underline{0.63$\pm$0.08}&\textbf{0.88$\pm$0.17}& 0.58$\pm$ 0.02& 0.11$\pm$ 0.04&0.37$\pm$ 0.03\\
{Middle}&{0.65$\pm$0.01}&\textbf{0.88$\pm$0.02}&{0.55$\pm$ 0.01}& 0.45$\pm$ 0.00&\underline{0.73$\pm$ 0.02}\\
{HandOut}&\textbf{0.89$\pm$0.01}&\underline{0.88$\pm$0.02}& 0.57$\pm$ 0.01& 0.59$\pm$ 0.01&  {0.76$\pm$ 0.03}\\
{Toe2}&{\bf 0.84$\pm$0.02}&\underline{0.78$\pm$0.00}& 0.60$\pm$ 0.01& 0.58$\pm$ 0.12&0.71$\pm$ 0.04\\
{Strawbry}&{0.44$\pm$0.03}&\textbf{0.95$\pm$0.01}&\underline{0.56$\pm$ 0.01}&0.54$\pm$ 0.00& 0.53$\pm$ 0.02\\
{ECG5000}&\textbf{0.92$\pm$0.03}&\underline{0.90$\pm$0.01}&0.59$\pm$0.02&{0.69$\pm$ 0.00}& 0.58$\pm$ 0.01\\
{Wafer}&{\bf 0.73$\pm$0.01}&0.31$\pm$0.04 & \underline{0.57$\pm$0.01}&{0.54$\pm$ 0.01}& 0.52$\pm$ 0.01\\
{Distal}&\underline{0.79$\pm$0.01}&{\bf 0.89$\pm$0.05}&0.58$\pm$0.03& 0.63$\pm$ 0.01&{0.78$\pm$0.02}\\
{Proximal}& {0.52$\pm$0.01}&{\bf 0.89$\pm$0.02} &{0.59$\pm$0.02}& 0.27$\pm$ 0.00&\underline{0.77$\pm$0.05}\\
{Phalanges}&{0.66$\pm$0.01}&{\bf 0.87$\pm$0.01}&{0.58$\pm$0.01}& 0.45$\pm$ 0.00&\underline{0.79$\pm$ 0.02}\\
{Dodger}&\underline{0.93$\pm$0.01}&{0.80$\pm$0.01}&0.69$\pm$0.02& 0.48$\pm$ 0.01&\textbf{0.97$\pm$0.00}\\
{Earthqks}&{0.47$\pm$0.01}&\textbf{0.73$\pm$0.03} &0.51$\pm$0.06& 0.45$\pm$ 0.05&\underline{0.54$\pm$0.08}\\
{ECG200}&\underline{0.71$\pm$0.03}&\textbf{0.87$\pm$0.01}&0.60$\pm$0.05& 0.63$\pm$ 0.06&{0.63$\pm$0.06}\\
 \hline
\multicolumn{6}{|c|}{\cellcolor{gray!10}{{\bf Reciprocal Rank (RR) scores}}}\\
\hline
Dataset&{WaRTEm-AD}&DTW \hspace{0.5cm}&ANN & BeatGAN & LOF \hspace{0.5cm}\\
&& &&& \\
\hline
{Toe1}&{0.66$\pm$.28}&{\bf 1.00$\pm$0.00}&\underline{0.83$\pm$0.29}& 0.41$\pm$0.12&\underline{0.83$\pm$0.28}\\
{Synthetic}&0.35$\pm$0.56&\textbf{0.67$\pm$0.57}&\underline{0.39$\pm$0.53}& 0.04$\pm$0.02&0.11$\pm$0.04\\
{Middle}&\textbf{1.00$\pm$0.00}&{\bf 1.00$\pm$0.00}&\underline{0.66$\pm$0.29}& 0.07$\pm$0.35&\underline{0.66$\pm$0.28}\\
{HandOut}&\textbf{1.00$\pm$0.00}&\textbf{1.00$\pm$0.00}&\textbf{1.00$\pm$0.00}&\textbf{1.00$\pm$0.00}&\textbf{1.00$\pm$0.43}\\
{Toe2}&0.30$\pm$0.05&1.00$\pm$0.09&\underline{0.50$\pm$0.43}&{\bf 0.66$\pm$0.47}&\underline{0.50$\pm$0.54}\\
{Strawbry}&0.05$\pm$0.04&0.09$\pm$0.04&\underline{0.13$\pm$0.11}&0.01$\pm$0.00&\textbf{0.37$\pm$0.02}\\
{ECG5000}&{\bf 1.00$\pm$0.00}&{\bf 1.00$\pm$0.00}&0.17$\pm$0.03&\textbf{1.00$\pm$0.00}& 0.17$\pm$0.28\\
{Wafer}&\textbf{1.00$\pm$0.00}&0.01$\pm$0.01&\underline{0.37$\pm$0.55}&0.16$\pm$0.01&0.05$\pm$0.28\\
{Distal}&\textbf{1.00$\pm$0.00}&1.00$\pm$0.29&\textbf{1.00$\pm$0.00}&\textbf{1.00$\pm$0.00}&\underline{0.83$\pm$0.28}\\
{Proximal}&0.33$\pm$0.00&{\bf 0.83$\pm$0.51}&{0.35$\pm$0.00}& 0.01$\pm$0.09&\underline{0.66$\pm$0.00}\\
{Phalanges}&\textbf{1.00$\pm$0.00}&\textbf{1.00$\pm$0.29}&\textbf{1.00$\pm$0.27}&\textbf{1.00$\pm$0.00}&\textbf{1.00$\pm$0.28}\\
{Dodger}&0.25$\pm$0.00&0.66$\pm$0.10&0.51$\pm$0.49&\textbf{1.00$\pm$0.00}&\underline{0.83$\pm$0.03}\\
{Earthqks}&0.02$\pm$0.00&{\bf 0.44$\pm$0.23}&0.09$\pm$0.05&\underline{0.42$\pm$0.12}&0.10$\pm$0.03\\
{ECG200}&0.50$\pm$0.00 &0.27$\pm$0.05&0.55$\pm$0.42&\underline{0.62$\pm$0.53}&{\bf 0.66$\pm$0.28}\\
\hline
\hline
\end{tabular}
}
\label{tab:seqresults}
\end{center}
\end{table}
\subsection{Sequence Anomaly Detection}
{\bf Datasets:} In our sequence anomaly evaluation, we use several datasets from UCR repository~\citep{UCRArchive2018} that have one class described as normal and the other as abnormal. This normal/abnormal classification suits the anomaly detection semantics. For example, in case of ECG datasets, one class is normal heart beat and other class indicates some case of heart failure. Similarly for ToeSegmentation dataset and HandOutline dataset one class shows normal walk, outline of a normal hand and other classes indicate abnormal walk, outline of defective hand respectively. 

\noindent{\bf Baseline Methods:} There is no method which targets the task of warping-robust sequence anomaly detection; thus, we use the following methods as our baseline methods:
\begin{itemize}
    \item {\bf ANN}: This is an auto-encoder model  with the same architecture as that of the auto-encoder part of  WaRTEm-AD. The representation size and the calculation of $AS(\cdot)$ with the help of Euclidean distance in the embedding space identical to WaRTEm-AD.
    \item {\bf DTW}: In this case, the $AS(\cdot)$ for a time sequence is calculated using $K$-$NN$  method considering DTW (dynamic time warping) distances between time series sequences. We set the DTW warping window size parameter to $10\%$ of sequence length, making it comparable to the choice of extent of warping operations in WaRTEm-AD. 
    \item {\bf BeGAN-U}: BeatGAN \citep{beatgan} model trained in an unsupervised way with the whole data-set (anomalous + non-anomalous points) forms another baseline. 
     \item {\bf LOF}: $LOF$\citep{breunig2000lof} method, considers each time sequence as a multidimensional datapoint, and defines anomaly score based on neighbor density.
\end{itemize}


\subsubsection{{\bf Results and Analysis}}
Table \ref{tab:seqresults} shows the comparative evaluation on PR-AUC, ROC-AUC and RR values obtained for each method, averaged across random initializations. 
It can be observed from the results that WaRTEm-AD performs better for the vast majority of datasets achieving the top-place in six out of fourteen datasets, and the second place in four of the remaining eight, on PR-AUC metric. The margin of improvement given by WaRTEm-AD method compared to other baselines is substantial with range varying from $10\%$ to $25\%$ where it is the top-performer, and for rest of the cases WaRTEm-AD method performance is reasonably close to the best performing method.
DTW is seen to be our closest competitor, with LOF being next. DTW method is explicitly designed for warping robustness.  The trends observed for PR-AUC, holds reasonably well for ROC-AUC also.\\
From statistical significance t-test(at $p<0.05$), it was observed that WaRTEm-AD is statistically significant to all other baseline methods, except DTW and LOF, as observed from the AUC metrics. 

\subsubsection{{\bf Sensitivity Analysis}} We analysed the sensitivity to the neighborhood parameter $K$ and warp operators on PR-AUC and ROC-AUC values. It was observed that the interpolation operator was somewhat more effective than the copy operator, in detecting sequence anomalies. The standard deviations in Table~\ref{tab:seqresults} show that the performance of the proposed method is robust to random initialization and is not highly sensitive to varying $K$ for almost all datasets. We found maximum gain in AUC by changing K in a data set is around $0.02$, indicating stability on $K$. 
 


\begin{table}
\caption{{\bf Sub-Sequence Anomaly Evaluation:} Comparison of ROC-AUC, PR-AUC and RR values for  WaRTEm-AD, DTW, ANN, MERLIN, BeatGAN (unsupervised version) and LOF. Each element of table shows mean values and std for each dataset over various K values for each of the metrics compared} 
\begin{center}
\resizebox{\textwidth}{!}{
\begin{tabular}{|l|cccccc|}
\hline
\hline
\multicolumn{7}{|c|}{\cellcolor{gray!10}{{\bf PR-AUC}}}\\
\hline
Dataset&{WaRTEm-AD}&  \hspace{0.5cm} DTW \hspace{0.5cm} &ANN &MERLIN& BeatGAN &LOF\\
&&  &&&\small{(unsupervised)}&\\
\hline
{dutch\_power\_demand}&\underline{ 0.35 $\pm$0.01}&{\bf 0.41 $\pm$0.04}& 0.17 $\pm$0.03&\cellcolor[gray]{0.8}& -& 0.36$\pm$0.06\\
{Marotta\_Valve\_Tek14}&{\bf 0.75$\pm$0.03}&\underline{0.71$\pm$0.12}& 0.28 $\pm$0.13&\cellcolor[gray]{0.8}& {0.15 $\pm$ 0.00}& 0.24$\pm$ 0.05\\
{Marotta\_Valve\_Tek16}&{\bf 1.00 $\pm$0.01}&{\bf 1.00$\pm$0.00}& 0.06 $\pm$0.02&\cellcolor[gray]{0.8}& \underline{0.18$\pm$0.00} & 0.10$\pm$0.04\\
{Marotta\_Valve\_Tek17}&{\bf 0.86 $\pm$0.02}& \underline{0.68 $\pm$0.45}& 0.25 $\pm$0.29&\cellcolor[gray]{0.8}&{0.17$\pm$0.00}& {0.15$\pm$0.12}\\
{chfdbch\_15}&{\bf 0.54 $\pm$0.64}&0.03 $\pm$0.01& 0.01 $\pm$0.00&\cellcolor[gray]{0.8}& \underline{0.06 $\pm$ 0.03}&{ 0.05$\pm$0.02}\\
{ann\_gun}&{\bf 0.93 $\pm$0.01}&\underline{0.46 $\pm$0.08}& 0.09 $\pm$ 0.04&\cellcolor[gray]{0.8}& -& 0.30$\pm$ 0.16\\
{Patient\_Respiration}&{ 0.15$\pm$0.00}&{\bf 0.52$\pm$0.00}& 0.05$\pm$0.01&\cellcolor[gray]{0.8}& {\bf 0.52$\pm$0.00}&\underline{ 0.35$\pm$0.27}\\
{Patient\_Respiration2}&{ 0.52$\pm$0.00}&\underline{0.53$\pm$0.01}& 0.04$\pm$0.01&\cellcolor[gray]{0.8}& {\bf 0.57$\pm$0.00}& \underline{0.53$\pm$0.01}\\
\hline
\multicolumn{7}{|c|}{\cellcolor{gray!10}{{\bf ROC-AUC}}}\\
\hline 
Dataset&{WaRTEm-AD}&  \hspace{0.5cm} DTW \hspace{0.5cm} &ANN & MERLIN& BeatGAN &LOF\\
&&  &&&\small{(unsupervised)}&\\
\hline
{dutch\_power\_demand}&{\bf 0.64 $\pm$0.04}&0.58 $\pm$0.01& 0.69 $\pm$0.08&\cellcolor[gray]{0.8}& -& \underline{0.61$\pm$0.16}\\
{Marotta\_Valve\_Tek14}&{\bf 0.88$\pm$0.02}&\underline{0.87$\pm$0.03}& 0.61 $\pm$0.04&\cellcolor[gray]{0.8}& 0.65 $\pm$ 0.01& 0.77$\pm$ 0.09\\
{Marotta\_Valve\_Tek16}&{\bf 1.00 $\pm$0.01}&\underline{0.98 $\pm$0.01}& 0.39 $\pm$0.09&\cellcolor[gray]{0.8}& 0.76 $\pm$0.00& 0.55$\pm$0.20\\
{Marotta\_Valve\_Tek17}&{\bf 0.98 $\pm$0.01}&\underline{0.97 $\pm$0.12}& 0.62 $\pm$0.40&\cellcolor[gray]{0.8}& 0.74 $\pm$0.00& 0.63$\pm$0.33\\
{chfdbch\_15}&{\bf 0.89 $\pm$0.04}&\underline{0.85 $\pm$0.13}& 0.49 $\pm$0.25&\cellcolor[gray]{0.8}& 0.84 $\pm$ 0.01& 0.83$\pm$0.09\\
{ann\_gun}&{\bf 0.97 $\pm$0.01}&\underline{0.92 $\pm$0.02}& 0.41 $\pm$ 0.09&\cellcolor[gray]{0.8}& - &0.70$\pm$ 0.15\\
{Patient\_Respiration}&{\bf 0.69$\pm$0.03}&0.55$\pm$0.04& 0.51$\pm$0.13&\cellcolor[gray]{0.8}& 0.59$\pm$0.00& \underline{0.68$\pm$0.09}\\
{Patient\_Respiration2}&0.68$\pm$0.04&0.56$\pm$0.03& 0.54$\pm$0.07&\cellcolor[gray]{0.8}& {\bf 0.89$\pm$0.00}& \underline{0.73$\pm$0.01}\\
\hline
\multicolumn{7}{|c|}{\cellcolor{gray!10}{{\bf Reciprocal Rank (RR) scores}}}\\
\hline
Dataset&{WaRTEm-AD}&  \hspace{0.5cm} DTW \hspace{0.5cm} &ANN&MERLIN  \hspace{0.5cm} & BeatGAN &LOF\\
&&  && &\small{(unsupervised)}&\\
\hline
{dutch\_power\_demand}&{\bf 1.00$\pm$0.00}&{\bf 1.00$\pm$0.00}& 0.25$\pm$0.00&\underline{0.33}&-&{\bf 1.00$\pm$0.00}\\
{Marotta\_Valve\_Tek14}&{\bf 1.00$\pm$0.00}&{\bf 1.00$\pm$0.00}&\underline{0.66$\pm$0.47}&0.00&0.11$\pm$0.01&0.36$\pm$0.17\\
{Marotta\_Valve\_Tek16}&{\bf 1.00$\pm$0.00}&{\bf 1.00$\pm$0.00}&0.10$\pm$0.02&0.00&0.14$\pm$0.00&\underline{0.21$\pm$0.06}\\
{Marotta\_Valve\_Tek17}&{\bf 1.00$\pm$0.00}&\underline{0.75$\pm$0.35}&0.54$\pm$0.65&0.13&0.18$\pm$0.00& 0.20$\pm$0.18\\
{chfdbch\_15}&\underline{0.13$\pm$0.04}&0.12$\pm$0.07&0.02$\pm$0.01&{\bf 1.00}&0.20$\pm$0.01&0.09$\pm$0.05\\
{ann\_gun}&{\bf 1.00$\pm$0.00}&{\bf 1.00$\pm$0.00}&0.08$\pm$0.06&{\bf 1.00}&-&\underline{0.66$\pm$0.47}\\
{Patient\_Respiration}&{ 0.50$\pm$0.00}&{\bf 1.00 $\pm$0.00}&0.10$\pm$0.03&0.33&{\bf 1.00$\pm$0.00}&\underline{0.75$\pm$0.35}\\
{Patient\_Respiration2}&{\bf 1.00$\pm$0.00}&{\bf 1.00$\pm$0.00}&0.06$\pm$0.03&0.00&{\bf 1.00$\pm$0.00}&\underline{0.60$\pm$0.00}\\
\hline
\hline
\end{tabular}
}
\end{center}
\label{tab:subresults}
\end{table}

\subsection{Sub-sequence Anomaly Detection}

{\bf Datasets:} We now perform an empirical evaluation over sub-sequence anomaly detection datasets viz., Space Shuttle Marotta Valve time series, Ann’s Gun dataset, BIDMC Congestive Heart Failure Database (record 15) and patient respiration dataset~\citep{lstmad,hotsax,series2graph,hu2019novel,senin2015time}.
 
\noindent{\bf Baselines:} We use the same baselines as in  sequence anomaly evaluation. Additionally, we also compare against MERLIN which can identify sub-sequence discords; we set sub-sequence range to 5-150. \\

\noindent{\bf Results and Analysis :} The sub-sequence empirical results are in  Table~\ref{tab:subresults}. It can be observed that WaRTEm-AD method outperforms all baseline methods compared identifying exact anomaly sequences with very few false alarms.  DTW is the closest competitor to WaRTEm-AD which is expected. False positive rate in BeGAN-U has been observed high. We could not get results for BeGAN-U for two data sets, as sequence reconstruction error was higher than specified threshold. ANN works well when anomaly is due to extreme values. LOF fails to identify subtle anomalies as in case of MarottaValve16 dataset as we will see soon. Table~\ref{tab:subresults} does  not have MERLIN results on PR-AUC and ROC-AUC metrics, since its design of returning just the top anomaly makes it unsuitable for those, where all methods were analyzed on the Reciprocal Rank metric (Ref. Section~\ref{pointevaluation})




\begin{wrapfigure}[12]{r}{0.5\linewidth}
         \vspace{-1cm}
         \centering
         \includegraphics[width=0.48\textwidth]{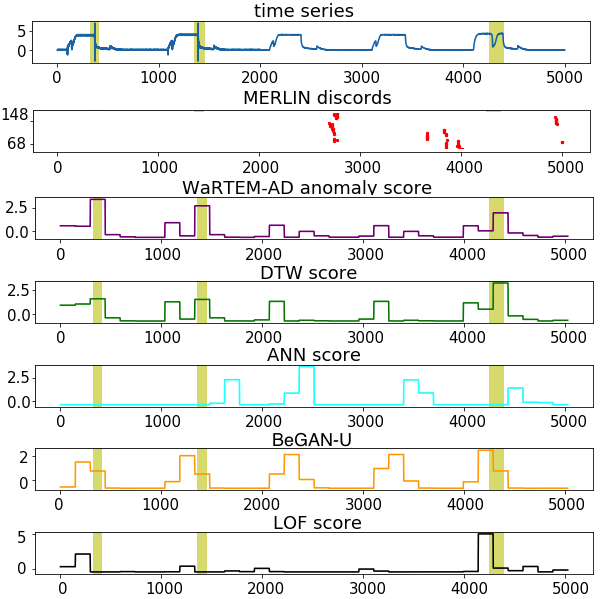}
          \caption{ MarottaValve16 time series with anomaly annotations \& scorings.}
        \label{fig:marotta17}
\end{wrapfigure}
As an example, we show Marotta Valve time series along with anomaly scorings obtained from various methods in Fig. \ref{fig:marotta17}. Among other methods, DTW raises some false alarms. LOF, on the other hand, captures longer flat part and shorter bump part (both can be considered as warp variant of other similar parts) as anomalies. ANN and MERLIN is seen to fail to detect anomalies for Marotta Valve datasets.

\subsection{Efficiency and Scalability}

We now discuss scalability of WaRTEm-AD on running times. While our task is unsupervised, the running time of the various methods may be seen as comprising two components; (i) training time that involves the model generation part, and (ii) inference time where the trained model is applied on each data point to derive anomaly scores. As expected, the former was seen to dominate the total running time. Fig.~\ref{fig:wgdseq} and Fig.~\ref{fig:wgd} show training time required for WaRTEm-AD, BeGAN-U and DTW on all sequence and sub-sequence anomaly detection datasets. Fig.~\ref{fig:wgdseq} and Fig.~\ref{fig:wgd} profile the training time on both the number of sequence in the dataset and length of each sequence, whereas Fig.~\ref{fig:trainpoint} illustrates the training time over length of time series for point anomaly task. All these plots illustrate the training efficiency of WaRTEm-AD, albeit being slower than BeGAN-U due to data augmentation.
\begin{wrapfigure}[11]{r}{0.30\linewidth}
       \vspace{-0.7cm}
         \centering
         \includegraphics[width=0.99\linewidth]{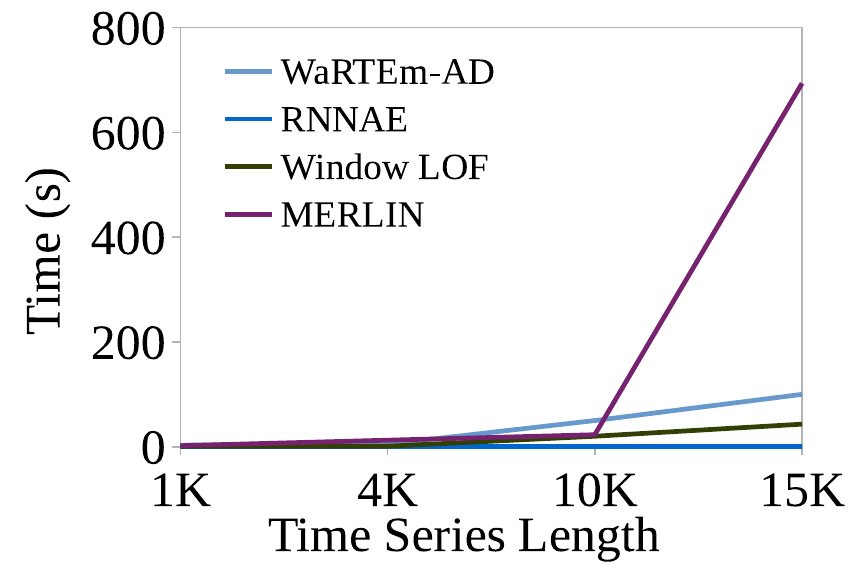}
             \caption{Inference time for point anomaly detection}
             \label{fig:infpoint}
\end{wrapfigure}
Fig.~\ref{fig:infpoint} illustrate the inference time on point anomaly tasks against time series length;  MERLIN is slow at inference time, all other methods are able to complete inferencing within $3$-$20$ seconds.
The sequence and sub-sequence anomaly inference times for all methods are comparable expect for  DTW.
These results illustrate the high scalability, both in training and inference time, achieved by WaRTEm-AD. This establishes that WaRTEm-AD is a method that is suitable for usage over large datasets in real-world scenarios.


\begin{figure*}[htp]
   \subfloat[Time for WaRTEm-AD and RNNAE for point anomaly datasets ]{\label{fig:trainpoint}
      \includegraphics[width=0.28\textwidth]{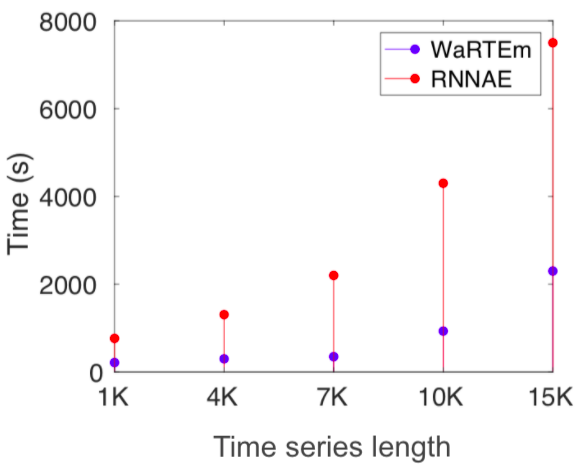}}
~
   \subfloat[Time for WaRTEm-AD, BeGAN-U and DTW methods on sequence anomaly datasets]{\label{fig:wgdseq}
      \includegraphics[width=0.35\textwidth]{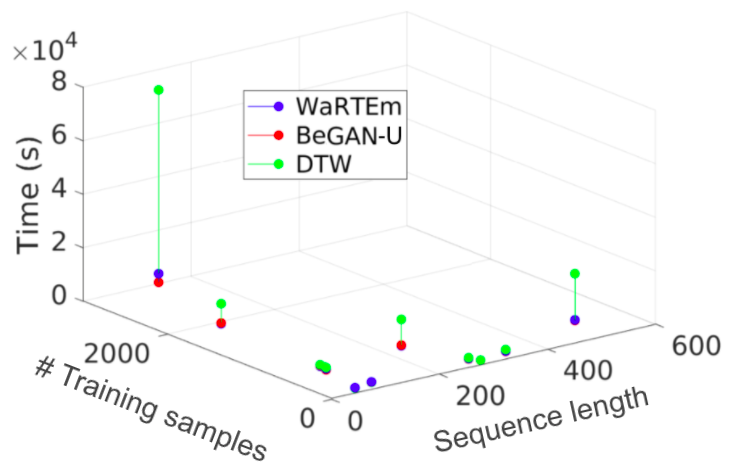}}
~      
    \subfloat[Time for WaRTEm-AD, BeGAN-U and DTW methods on sub-sequence anomaly datasets]{\label{fig:wgd}
      \includegraphics[width=0.32\textwidth]{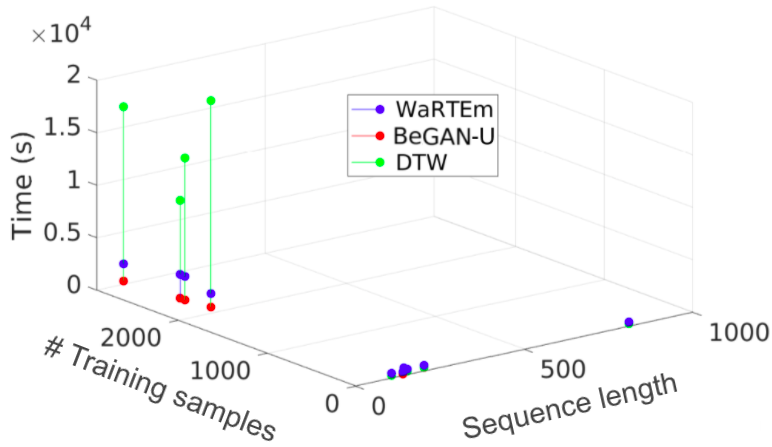}}
   \caption{Training time analysis}
   \label{fig:timeanalysis}
\end{figure*}

\section{Conclusion and Future Work}

In this paper, we considered the problem of time series anomaly detection, and outlined the importance of distinguishing warping from other kinds of variations in time series, within the tasks. We devised an anomaly detection framework, { WaRTEm-AD}, which can be used for both sequence and point anomaly detection tasks, making it a general purpose building block for time series anomaly detection. {WaRTEm-AD} employs a two-phase approach, with the first phase using a novel mechanism, that of { warping operators} for data augmentation, which is leveraged in a self-supervised learning framework to embed time series data into a vector space of pre-specified dimensionality. The second phase makes use of simple mechanisms to score anomalousness using local neighborhood statistics in line with the state-of-the-art in anomaly detection over non-temporal data. Through an extensive set of experiments over real-world data, we established the empirical effectiveness of {WaRTEm-AD} on both tasks, with it comparing well or outperforming the state-of-the-art in most point and sequence/sub-sequence anomaly detection scenarios. Code is available at https://github.com/WaRTEm-AD/UnivariateAnomalydetection.\\

\noindent{\bf Future Work:} In future, we are interested to make { WaRTEm-AD} model robust to other kinds of legitimate time series distortions within specific application domains. Our goal is to develop a general-purpose framework that is flexible enough to be tuned for robustness to user-specified kind of noises, to achieve applicability across a variety of domains. We are also considering utility of {WaRTEm-AD} in anomaly detection over multi-variate time series data, and exploring ways of devising clever heuristics towards combining the {\it copy} and {\it inter} methods. We are also exploring the usage of { WaRTEm-AD} in astronomical data analysis where there is an interest towards identifying causation of outlying events using temporal and ordering cues. 


\backmatter





\bmhead{Acknowledgments}
The work is supported by project titled “Robust Multi-view Learning for Extreme Events Detection and Prediction in Time Series Data”, IITPKD/2021/013/CSE/SAB funded by ICSR

\bibliography{ref}

\end{document}